\documentclass[letterpaper, 10 pt, conference]{format/ieeeconf}
\usepackage{amsmath}
\usepackage{amsfonts}
\usepackage{setspace}
\usepackage[space, compress, sort]{cite}

\usepackage{amsthm}
\usepackage{amssymb,stmaryrd}
\usepackage[algo2e]{algorithm2e} 
\usepackage{mathtools}
\usepackage{tabularx}
\usepackage{graphicx}
\usepackage{subfigure}
\usepackage{enumerate}
\usepackage{float}
\usepackage{url}
\usepackage{verbatim}
\usepackage{booktabs} 
\usepackage{multirow}
\usepackage[linkcolor=black,citecolor=black,urlcolor=black,colorlinks=true]{hyperref}
\usepackage{bm}
\usepackage{empheq}
\IEEEoverridecommandlockouts
\usepackage{graphicx}
\usepackage{algorithm}
\usepackage{algpseudocode}

\newtheorem{definition}{Definition}

\newcommand{\SB}[1]{{\color{black}{#1}}}

\begin{document}

\title{Multi-Agent Exploration of an Unknown Sparse Landmark Complex via Deep Reinforcement Learning }

%Multi-Agent Exploration of Landmark Complex in Sparse Landmark Environments via Deep Reinforcement Learning (TBD)
\author{Xiatao Sun$^{1}$, Yuwei Wu$^{1}$, Subhrajit Bhattacharya$^{2}$, Vijay Kumar$^{1}$% <-this % stops a space
\thanks{This work was supported in part by ARL DCIST CRA under Grant W911NF-17-2-0181, and in part by NSF under Grants CCR-2112665.}
\thanks{$^{1}$ Xiatao Sun, Yuwei Wu, and Vijay Kumar are with the GRASP Laboratory, University of Pennsylvania, Philadelphia,
PA 19104, USA
        {\tt\small \{sxt, yuweiwu, kumar\}@seas.upenn.edu}}%
\thanks{$^{2}$ Subhrajit Bhattacharya is with the Department of Mechanical Engineering and Mechanics, Lehigh University, Bethlehem, PA 18015, USA
        {\tt\small 	sub216@lehigh.edu}}%
}

\maketitle
\thispagestyle{empty}
\pagestyle{empty}
%%%%%%%%%%%%%%%%%%%%%%%%%%%%%%%%%%%%%%%%%%%%%%%%%%%%%%%%%%%%%%%%%%%%%%%%%%%%%%%
%layout: 1. generate exploration under the unknown environment is hard and challenged and the mainstream rely on localization and metric. 2. the complex based representation don't rely on such kind of information, while need some prior design for the environment. 3. prior work have strong assumption and limits, while lacking communication strategies, cannot work well under destruction. 4. our work can robustly handle such situation using DRL 5. detailed method used. 5. performance

\begin{abstract}

% Landmarks in a GPS-denied environment can aid a swarm of robots in exploring and generating a landmark complex, a simplicial complex constructed from observation of landmarks without localization and metric. Prior works on landmark complex exploration are under the strong assumption of sufficient landmarks. Therefore, prior methods are susceptible to landmark reduction. If some landmarks are destroyed and missing in the environment, the dispersion of landmarks will become larger, and this assumption will no longer hold. This paper proposes the first use of deep reinforcement learning for multi-robot cooperative exploration with limited sensing capability in sparse landmark environments while reducing communication. We combine individual and group rewards to encourage a collaborative exploration policy. We design a three-stage curriculum to mitigate the group reward sparsity and speed up training by gradually adding random obstacles and destroying random landmarks. Simulation experiments show that the final learned policy has over 20\% improvement compared with the state-of-the-art landmark complex exploration method in sparse landmark environments.

\SB{In recent years}
Landmark Complex\SB{es} \SB{have been successfully employed for} localization-free and metric-free autonomous exploration using a group of sensing-limited and communication-limited robots in a GPS-denied environment. To ensure %the performance of the 
rapid \SB{and complete} exploration, existing works 
% usually rely on the necessary 
\SB{make}
assumptions on the \SB{density and} distribution \SB{of landmarks in the environment.} These assumptions may be overly restrictive\SB{, especially in} hazardous environments where landmarks \SB{may be} destroyed \SB{or} completely missing. In this paper, we first propose a deep reinforcement learning framework for multi-agent cooperative exploration in environments with sparse landmarks while reducing client-server communication.
By leveraging recent development on partial observability and credit assignment, our framework can train the exploration policy efficiently for multi-robot systems. The policy receives individual rewards from actions based on a proximity sensor with limited range and resolution, which is combined with group rewards to encourage collaborative exploration \SB{and construction of the Landmark Complex through observation of 0-, 1- and 2-dimensional simplices}. In addition, we employ a three-stage curriculum learning strategy to mitigate the reward sparsity by gradually adding random obstacles and destroying random landmarks. Experiments in simulation demonstrate that our method outperforms the state-of-the-art landmark complex exploration method in efficiency among different environments with sparse landmarks.
%looks like they mention the GPS-denied condition: consider the problem of autonomous exploration of an unknown, GPS-denied environment using a swarm of robots with very limited resources and limited sensing capabilities. 

\end{abstract}
%%%%%%%%%%%%%%%%%%%%%%%%%%%%%%%%%%%%%%%%%%%%%%%%%%%%%%%%%%%%%%%%%%%%%%%%%%%%%%%%
\section{Introduction}

%%%%%%%%%%%part 1. Big problem in science: use a paragraph to explain the mainstream exploration techs

%%%%%%%%%%%%%%%%%%%%%%%%%%%%%%%%%%%%%%%%%%%%%%%%%%%%%%%%%%%%%%%%%%%%%

% The robots have no a priori knowledge of the environment and need to rapidly explore and construct a map in a distributed manner using existing landmarks, the presence of which can be detected using onboard senors, although little to no metric information (distance or bearing to the landmarks) is available.

% presence of a necessary density/distribution of landmarks is ensured by design of the urban/indoor environment.
%%%%%%%%%%%%%%%%%%%%%%%%%%%%%%%%%%%%%%%%%%%%%%%%%%%%%%%%%%%%%%%%%%%%%

% as an exp: 
% traditional/mainstream methods use semantics/features/..frontiers for exploration, however it rely on odometry/coordinates, and demand computation for vio systems, as some work 1, 2, 3, ...., and also very difficult and expensive for real world scenarios (for example, the fire case, it's not worthy to use such a expensive swarm)..... For a group ofhttps://www.overleaf.com/project/62c4aa4180e4074d1e2630a6 limited resources robots... landmark based method can efficiently ... do something with designed environment that landmarks are well distribution. 
Exploration \SB{and mapping} of unknown \SB{environments} is one of the crucial applications for mobile robots. The \SB{existing} approaches mostly \SB{fall under the} simultaneous localization and mapping (SLAM) \SB{literature}, which focuses on constructing different environment representations, such as metric \cite{folkesson2005slam, pan2021slam}, topometric \cite{blanco2007slam, min2012slam}, or topological maps \cite{blochliger2018slam, xue2020slam}. However, many rely on expensive sensors for high-quality measurement and intense computation for odometry estimation and error correction to ensure \SB{good} map quality, which becomes more difficult and unaffordable for a large-scale multi-agent exploration, especially in hazardous environments where robots can easily get damaged or disconnected.

%%%%%%%%%%% part 2. Narrower problem within: our problem definition
The appearance of \textit{Landmark Complex}\cite{ghrist2012topological}, a topological representation, provides a new way to aid robots with limited resources and capabilities in exploring an unknown, GPS-denied environment. Besides the inherent advantages of topological maps, which are less computationally intensive, and more geometrically and semasiologically informative \cite{kostavelis2015semantic},  the landmark complex can also improve the efficiency and scalability of cooperative multi-agent exploration as its lightweight information sharing and coordinate-free navigation. Autonomous exploration with existing landmarks using landmark complex representation has been validated among different environments and extended to multiple robots\cite{teymouri2021landmark}. However, harsh environmental conditions make landmarks susceptible to destruction in hazardous environments, which is undesired for search and rescue missions and supervision. For example, visual landmarks, such as AprilTags and semantic objects, are possibly burnt by fire or destroyed by an earthquake; signal landmarks, such as wireless routers and 5G antennae, can malfunction when facing electromagnetic pulses caused by nuclear disasters. Without harsh environmental conditions, landmarks can still be damaged due to aging after deployment.

\begin{figure}[t]
      \centering
      \includegraphics[scale=0.31]{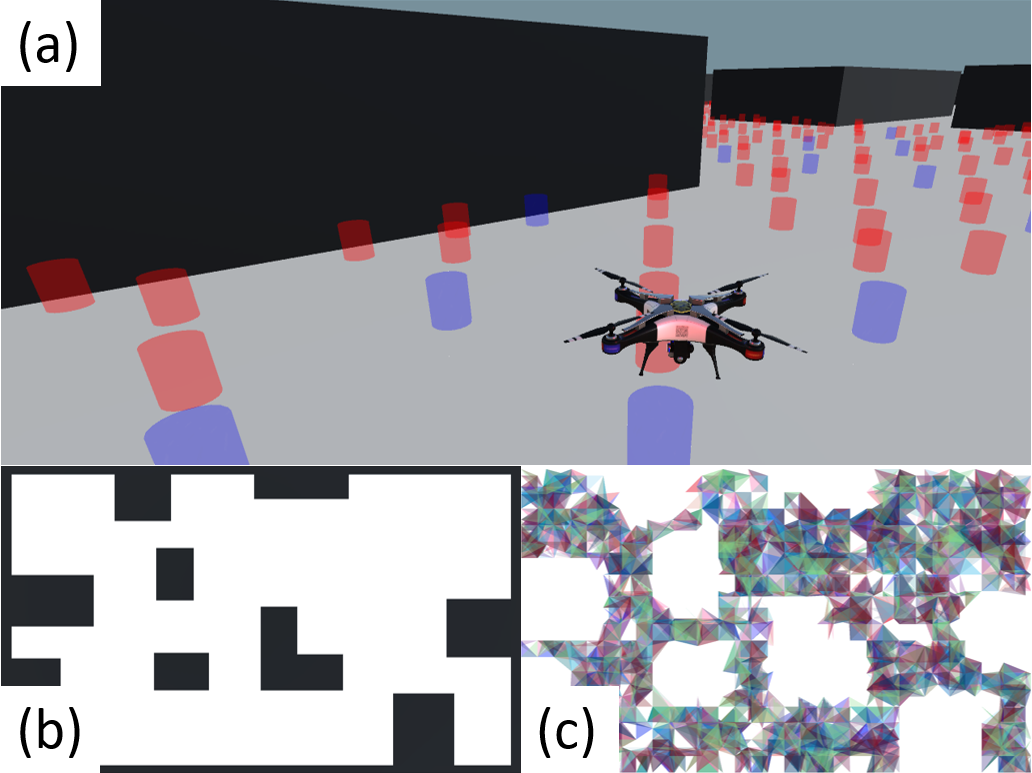}
      \caption{(a) The agent explores an environment with detectable landmarks (red) and destroyed landmarks (blue). (b) Overview of the environment with randomly generated obstacles. (c) Landmark complex constructed in the environment with sparse landmarks. The complementary video can be accessed at the following link: 
\url{{https://youtu.be/8ujQVo_YvF0}}.}
      \label{overview_illustration}
      \vspace{-0.6cm}
\end{figure}

%%%%%%%%%%% part 3. Yet narrower paper Gap: the issues of previous work

%%%%%%%%%%%%%%%%%% the traditional landmark based exploration, as they did not consider the sparse landmark
 \SB{Existing} landmark complex-based exploration approaches assume a necessary density of landmarks in an environment. To guarantee this assumption, they either decrease the dispersion of landmarks by placing more landmarks \cite{ramaithititima2018landmark} or use a strategic method that filters the environment over a list of sensor footprints and places landmarks in uncovered areas \cite{teymouri2021landmark}. No matter which landmark placement algorithm is applied, previous landmark complex exploration strategies heavily depend on the assumption of sufficiently dense landmarks. Moreover, the existing approaches require constant client-server communication, resulting in packet loss in communication-limited hazardous environments.

%%%%%%%%%%% part 3. Our work summary: In order to bridge this gap, we .....

% Our work seeks to push the landmark complex further toward real-world scenarios by considering landmark reduction and reducing client-server communication. Leveraging the recent development on partial observability and credit assignment in multi-agent deep reinforcement learning (MADRL) \cite{wong2021multiagent}, we formulate our objectives into a MADRL problem. Although MADRL has been applied to various exploration problems and demonstrated better performance compared with human-engineered methods \cite{geng2018learning, he2020decentralized,yu2021learning}, to the best of our knowledge, this is the first work using MADRL to landmark complex exploration.
% To address the issue of reward sparsity and speed up training, we take inspiration from the curriculum learning (CL) paradigm \cite{bengio2009curriculum} and design a three-stage curriculum for learning to explore in a sparse landmark environment.

To address the limitations above, the proposed approach aims to solve the problem of landmark-based exploration in more realistic scenarios with sparse landmarks and limited client-server communications using Multi-Agent Deep Reinforcement Learning (MADRL). By employing partial observability and credit assignment in Multi-Agent Posthumous Credit Assignment (MA-POCA) algorithm \cite{cohen2021use}, we propose Landmark Complex Curriculum with MA-POCA (L2C MA-POCA), a systematic framework for multi-agent landmark complex exploration. We take inspiration from the curriculum learning (CL) paradigm \cite{bengio2009curriculum} and design a three-stage curriculum for learning to explore in an environment with sparse landmarks. Our contributions are as follows:
% \begin{itemize}
%     \item An MADRL-based multi-robot exploration framework to adapt for landmark reduction and optimize client-server communication in landmark complex environments.
%     \item A three-stage curriculum to direct the training process and mitigate reward sparsity.
%     \item Benchmark tests in physics-based environments with randomly generated obstacles and randomly destroyed landmarks to show the robustness of our approach.
% \end{itemize}

\begin{itemize}
    \item An MADRL-based exploration framework to adapt for sparse landmarks with limited communications.
    \item A well-designed three-stage curriculum to guide the training process and mitigate reward sparsity.
    \item Analysis of various simulation results to show the learned policy outperforms the frontier-based method in environments with sparse landmarks.
\end{itemize}

\section{RELATED WORKS}

\subsection{Multi-agent Exploration with Limited Resources}

Most existing works on multi-agent exploration usually assume unlimited and perfect communication between robots \cite{julia2012planning}. However, real-world scenarios are often communication-restricted, which have not been well investigated \cite{amigoni2017multi}. Moreover, SLAM methods usually require high-performance sensors, such as LiDAR \cite{droeschel2018slam} and monocular \cite{artal2015slam} or stereo cameras \cite{engel2015slam}, which are not economically and technically scalable for a large system. Therefore, most current methods cannot be immediately extended to real-time multi-agent exploration if considering limited communication, hardware constraints, and inconsistency among global and local coordinate systems of different agents \cite{kshirsagar2018multislam}.

Some works add mechanisms to reduce communication frequency in well-studied exploration methods to save communication bandwidth. Rathnam \textit{et al.} \cite{rathnam2011multi} incorporate a utility function with a frontier-based algorithm to guarantee communication. Mahdoui \textit{et al.} \cite{mahdoui2018multi} propose a frontier-based method with frontier point exchanging between robots to reduce communication. Considering sensor limitations, others propose new topological representations for multi-agent exploration. Although some SLAM methods can also construct topological maps \cite{blochliger2018slam, xue2020slam}, they are still based on aggregated grids or voxels, which generate heavy information to be shared between robots. Zhang \textit{et al.} \cite{zhang2022topo} propose a new topological map constructed from places of observation. Ghrist \textit{et al.} \cite{ghrist2012topological} propose the landmark complex, a simplicial complex constructed from observed landmarks. Ramaithititima \textit{et al.} \cite{ramaithititima2018landmark} bring up the assumption of sufficient landmarks for landmark complex exploration and propose a corresponding multi-agent frontier-based exploration strategy under this assumption. Teymouri \textit{et al.} \cite{teymouri2021landmark} propose the Landmark Placement Algorithm (LPA) to improve the efficiency of landmark placement by filtration over a list of sensor footprints. In this work, Teymouri \textit{et al.} also propose the Landmark Complex Construction Algorithm (LCCA), a more refined multi-agent exploration strategy based on LPA. However, like prior works in landmark complex exploration, LCCA heavily relies on dense landmark distribution while does not consider the communication burdens. Hence, it is still far from real-world applications. Our proposed approach aims at filling in these gaps.

\subsection{Reinforcement Learning and Curriculum Learning}

% MADRL
Recently, MADRL has been applied to various exploration problems and demonstrated better performance than traditional human-engineered methods, especially in complex and dynamic environments. Geng \textit{et al.} \cite{geng2018learning} \SB{applied} MADRL to exploring the environments as occupancy grids, while He \textit{et al.} \cite{he2020decentralized} \SB{applied} MADRL to exploring environments with structured patterns. Yu \textit{et al.} \cite{yu2021learning} propose an RL-based global planner for multi-agent active visual exploration using environmental relations and inter-agent interactions.

% CL
Modern reinforcement learning (RL) algorithms demand significant training steps to converge if under reward sparsity and without guidance \cite{eschmann2021reward}. CL, proposed by Bengio \textit{et al.} \cite{bengio2009curriculum} as a method to guide the training process from simple tasks to challenging tasks, can be used to guide the training process of RL. Luo \textit{et al.} \cite{luo2020accelerating} apply CL to speed up RL training for multi-goal reaching tasks and demonstrate improved training efficiency. Li \textit{et al.} \cite{li2020towards} propose an RL system to solve manipulation tasks and use CL to mitigate the utilization of sparse rewards. Chen \textit{et al.} \cite{chen2019end} apply CL to RL-based multi-agent exploration with visual agents and demonstrate better performance than the state-of-the-art frontier-based method. 

Our work follows a similar strategy to \cite{chen2019end}. In contrast to this work, our approach can perform topological exploration with limited-range sensors in significantly larger physics-based environments instead of in a 2-dimensional \(20\times20\) grid world with global visual observation.

\section{Preliminaries}

\subsection{Landmark Complex}

%% use some mathematical language to restate and formulate this problem

%(1) define the  simplicial complex, you can write again about these math formulation.
\SB{We start with the definition of a general simplicial complex, which is a natural higher dimensional extension of graphs, and of which Landmark Complex is a specific type.

\begin{definition}[A $m$-simplex over $V$]
Given a set $V$ (the elements of which are referred to as \emph{points}), a \emph{$m$-dimensional simplex} (or simply a \emph{$m$-simplex}) is a subset of $V$ with $(m+1)$ elements.
\end{definition}
}

\begin{definition}[A Simplicial Complex~\cite{hatcher2002algebraic} over $V$]
A simplicial complex, $K$, 
% which is the foundation of a landmark complex, 
\SB{over the set $V$}
is a collection $K = \{C_0, C_1, C_2, ...\}$,
% constructed over a vertex set \(V\) 
\SB{such that 
% $C_m$ is a collection of $m$-simplices with the following properties:
}
\begin{enumerate}[i.]
    \item 
    % Every element \(\sigma \in K\) is an \(m\)-simplex, \(m = 0, 1, 2, ...\), and a subset of \(V\) with cardinality of \(m+1\).
    \SB{$C_m$ is a collection of $m$-simplices over $V$, and thus $\sigma\in C_m$ is a $m$-simplex ({i.e.}, $\sigma\subseteq V$ with $|\sigma| = m+1$)}
    \item \SB{For all} \(m \geq 1\), \SB{if $\sigma\in C_m$ and} \(v \in \sigma\), \SB{then} \(\sigma - v ~ \SB{\in C_{m-1}}\) \SB{and is called} a \emph{face} of the simplex \(\sigma\).
\end{enumerate}
 \end{definition}
 
%(2) landmark complex definition, and the 
A landmark complex, \(\mathcal{X} = \{C_0, C_1, C_2, ...\}\), is an abstract simplicial complex \SB{over a set} of identifiable landmarks, \(\mathcal{L} = \{L_1, L_2, L_3,...\}\), \SB{constructed} from observations and is designed to help multi-agent systems with limited sensors construct a correct topological map of an environment \cite{ghrist2012topological}. For every \(l\)-tuple of landmarks that \SB{are} inside the sensor footprint during \SB{an} observation, the corresponding \((l - 1)\)-simplex is inserted to \(C_{l - 1}\), \SB{and all} its faces and sub-faces are also inserted in \(C_i\), \(i ~\SB{<}~ l - 1\).

\begin{figure}[!t]
      \centering
      %\vspace{-0.5cm}
      \includegraphics[scale=0.255]{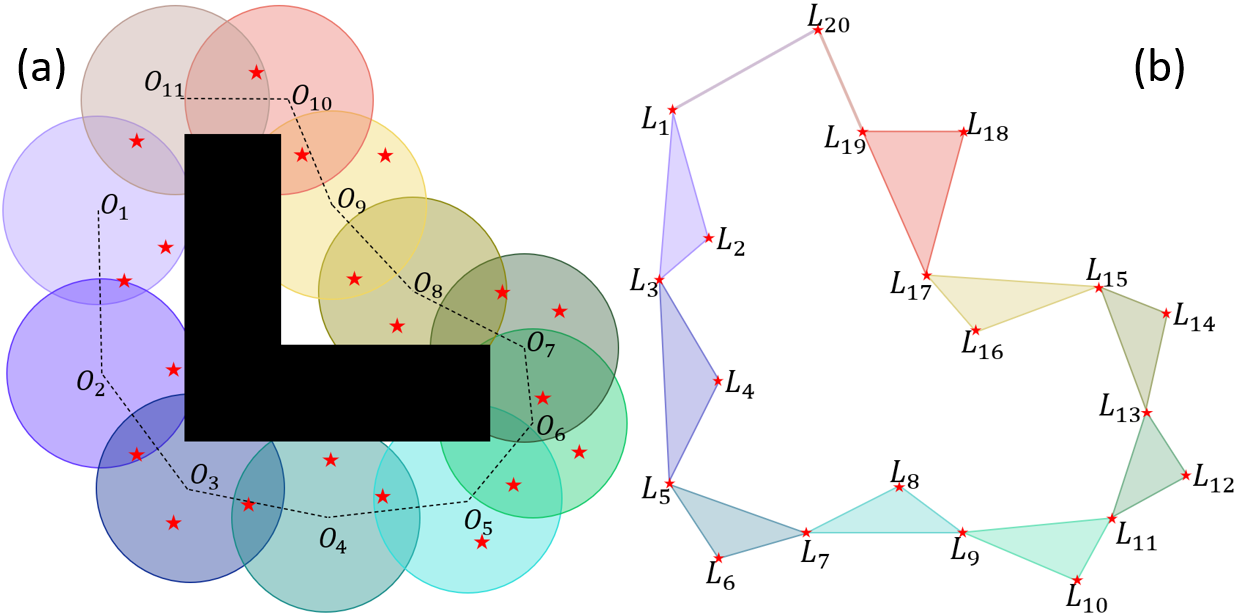}
      \caption{A robot in an environment (a) observes landmarks (red stars) and constructs a landmark complex (b) from them. The landmark complex is a topological representation of the environment.}
      \label{landmark_complex_illustration}
      \vspace{-0.6cm}
\end{figure}

Figure \ref{landmark_complex_illustration} illustrates an example of a landmark complex constructed from 11 observations with an omni-directional proximity sensor. At the first observation point \(O_1\), landmarks \(\{L_1, L_2, L_3\}\) fall into the sensor footprint and are observed by the robot. These landmarks form a \(2\)-simplex (triangle), which is inserted to \(C_2\). Their lower dimensional counterparts, which are \(1\)-simplices (edges) \(\{\{L_1, L_2\}, \{L_1, L_3\}, \{L_2, L_3\}\}\) and \(0\)-simplices (nodes) \(\{\{L_1\}, \{L_2\}, \{L_3\}\}\) are inserted to \(C_1\) and \(C_0\) respectively. In this way, a landmark complex as shown on Figure \ref{landmark_complex_illustration}(b) can be constructed after 11 observations. A bidirectional graph \SB{constisting of the} \(0\)-simplices as nodes and \(1\)-simplices as edges can be constructed from this landmark complex for navigation \SB{and is referred to as the \emph{skeleton} of the complex}.
%sparsity of landmark distribution.
%like, the completed exploration environment is defined as the space filled with enough landmarks for exploration ... 

%In order to perform exploration in a landmark complex environment, the landmark distribution cannot be sparse. This sparsity can be quantified based on the concept of dispersion from sampling theory \cite{lavalle2006planning}:
%$$
%\delta(P) = \underset{x \in \chi}{sup}\{\underset{p \in P}{\min}\{\rho(x, p)\}\}
%$$
%where \(P\) is a finite set of samples in a metric space \((\chi, \rho)\).

%In \cite{ramaithititima2018landmark}, Ramaithititima \textit{et al.} use \(L_2\) metric to define the dispersion of the landmarks:
%$$
%\delta = \underset{x \in W}{sup}\{\underset{y_j \in Y}{\min}\{||x-y_j||\}\}
%$$
%which result in at least one landmark can be observed at every point in the environment.
To conduct exploration \SB{for constructing} a landmark complex \SB{representation of an} environment, most prior works assume that the landmark distribution cannot be too sparse in the environment. This assumption of sufficient landmarks is deduced by Ramaithititima \textit{et al.} based on sampling theory\cite{ramaithititima2018landmark}. The distribution of landmarks should ensure that at least one landmark is observable at every point in the environment, which can be satisfied using the LPA \cite{teymouri2021landmark} to place landmarks by filtration over a list of sensor footprints. % \(B = \{ B^1, B^2, B^3, ...\}\).

\subsection{Sparse Landmark Complex}
 % explain our defined problem: landmark fulfilled environment with  
Although the set of identifiable landmarks \(\mathcal{L}\) can be generated using LPA, the assumption of sufficient landmarks can be easily broken if robots operate in a hazardous environment where some landmarks are destroyed and cannot be observed during exploration. We define this problem of sparse landmarks as removing a set of randomly sampled landmarks \(\mathcal{L}_D\) from \(\mathcal{L}\). The remaining landmarks \(\mathcal{L}_R\) is the result of removing elements in \(\mathcal{L}_D\) from \(\mathcal{L}\) such that
$
\mathcal{L}_R := \mathcal{L} \backslash \mathcal{L}_D.
$
An environment with sparse landmarks can significantly reduce exploration efficiency since robots only detect the observed surrounding landmarks for localization-free and metric-free exploration. Sparse landmarks can also decrease the fidelity of the final topological map constructed from the landmark complex since the topology of the environment is captured by existing landmarks.

\section{Approach}
%%% change to your RL method, should we rename it? like a abbv to name you framework (Maybe L2C (Landmark Complex Curriculum)? or L2C MA-POCA)

Multi-agent cooperative exploration in a landmark complex environment can be defined as a Decentralized Partially Observable Markov Decision Process (Dec-POMDP) \cite{oliehoek2016pomdp} since each robot acts based on its local observation. This can be formalized as a tuple \((N, \mathcal{S}, \mathcal{O}, \mathcal{A}, P, \gamma, r)\), where \(N = \{1,...,n\}\) denotes the number of robots (agents), \(\mathcal{S}\) is the set of states in a landmark complex environment, \(\mathcal{O}\) is the set of joint observations, \(\mathcal{A}\) is the set of joint actions, \(P:\mathcal{S} \times \mathcal{A} \times \mathcal{S} \rightarrow [0,1]\) is the transition function, \(\gamma \in [0, 1]\) is the discount factor, and \(r:\mathcal{S} \times \mathcal{A} \rightarrow \mathbb{R}\) is the reward function. 
%Each agent has to make actions \(a\) based on belief states \(b\) since directly inferencing for real states \(s\) from local observations of landmarks \(o\) is not possible. 
Therefore, the problem of landmark complex exploration can be formulated as learning an optimal policy \(\pi^*(a|s;\theta_\pi)\) that maximizes the discounted expected cumulative reward under such Dec-POMDP setting
$$
\pi^*=\text{arg}\,\underset{\theta_\pi}{\text{max}} \, \mathbb{E} \left[\sum_{t=0}^{T-1} \gamma^t r(s_t, a_t) \right],
$$
where \(\theta_\pi\) denotes the neural network parameters that represent the policy, \(t\) denotes the current stage, and \(T\) represents the horizon of exploration. Therefore, instead of using frontier-based methods, we propose L2C MA-POCA by combining MA-POCA and curriculum to train an optimal policy under the Dec-POMDP setting. 

\begin{figure}[t]
      \centering
      \includegraphics[scale=0.29]{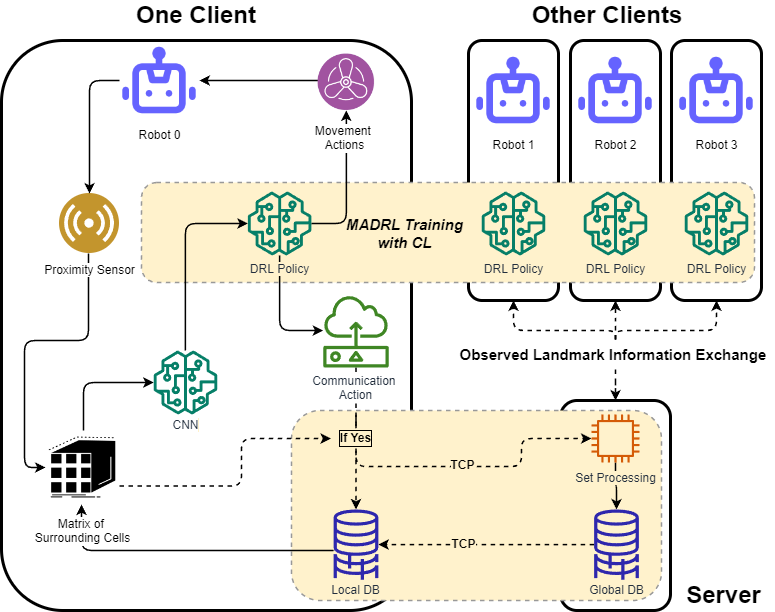}
      
      \caption{Overview of the proposed multi-agent landmark complex exploration framework via MADRL. The policy is trained using CL and deployed using a client-server architecture.}
      \label{framework_overview}
      \vspace{-0.6cm}
\end{figure}

% Framework overview and why use MA-POCA
As illustrated in Figure \ref{framework_overview}, L2C MA-POCA employs a client-server architecture similar to \cite{teymouri2021landmark} for synchronizing records of observed landmarks, while each agent performs inference and acts decentrally. The proposed framework applies MA-POCA in sparse landmark complex exploration. At the early stage of training, agents in the environment can generate a high volume of concurrent requests for updating their observations to the server, which can cause some agents to disconnect from the server. The disconnected agents experience early termination and create the credit assignment problem due to the non-steady size of inputs. This issue is further amplified if the time scale is increased during training. MA-POCA can efficiently handle the credit assignment problem using a self-attention mechanism, which can adapt to varying sizes of input better than absorbing states in other MADRL algorithms.

% Therefore, instead of using frontier-based methods, based on Multi-Agent Posthumous Credit Assignment (MA-POCA) \cite{cohen2021use}, we propose Landmark Complex Curriculum with MA-POCA (L2C MA-POCA), a novel landmark complex exploration framework using MADRL. As illustrated in Figure \ref{framework_overview}, L2C MA-POCA employs a client-server scheme similar to the prior work \cite{teymouri2021landmark}. But instead of moving based on the Dubins curve and detecting landmarks to its left or right, each robot is now holonomic and equipped with an omni-directional proximity sensor for several reasons. 
% First, the method in prior work is only supposed to work in pure numerical simulation, whereas one of our objectives is to push landmark complex exploration further towards practical usage by extending it to a physics-based environment, in which agents should be moved by applying force to change the acceleration rather than simply perform direct teleportation along the generated Dubins curve. Therefore, each agent is controlled by linear velocity on \(x\) and \(y\) axes ,\(v_x\) and \(v_y\), and angular velocity on \(z\) axis, \(\omega_z\). 
%Moreover, in a physics-based environment, we need to take collision into consideration.

% Agent setup (action)
Each client (agent) in L2C MA-POCA is holonomic and equipped with an omni-directional proximity sensor. Each agent moves by controlling \(v_x\) and \(v_y\), which are linear velocity on \(x\) and \(y\) axes, and \(\omega_z\), which is angular velocity on \(z\) axis. Besides the movement actions, each agent can also control whether to communicate with the server to send local observations of landmarks and receive incremental updates of observed landmarks for the local database or not. 

\begin{figure}[t]
      \centering
      \includegraphics[scale=0.265]{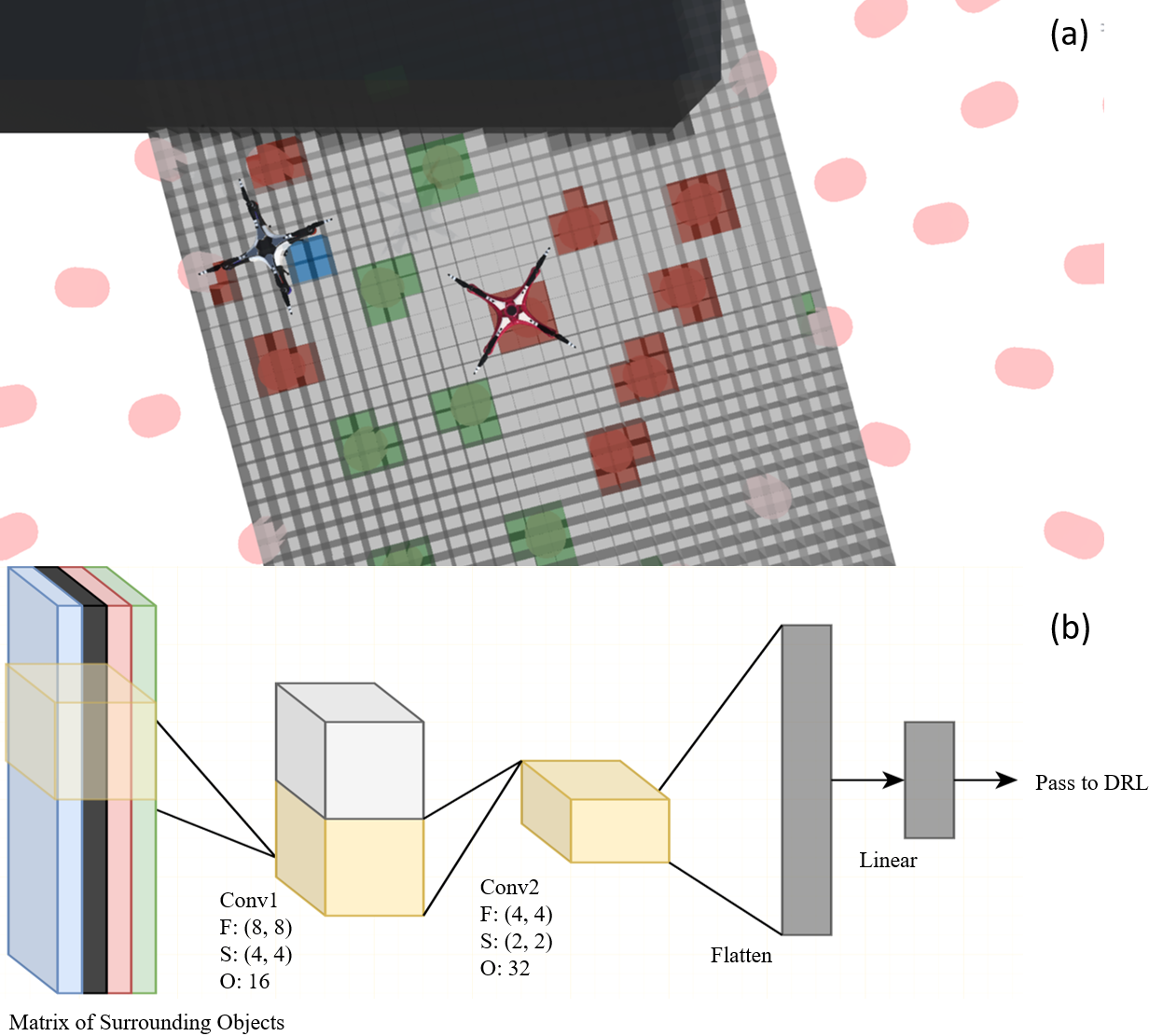}
      \caption{(a) The range-limited and resolution-limited omni-directional proximity sensor can detect and identify types of surrounding objects. The types include other agents, observed landmarks, unobserved landmarks, and obstacles, represented as blue, red, green, and dark grey cells respectively. (b) The CNN architecture for processing matrices that are output by the sensor. \emph{F}, \emph{S}, and \emph{O} represent filter size, stride, and output channel in the convolutional layers respectively.}
      \label{sensor}
      \vspace{-0.6cm}
\end{figure}

% Agent setup (observation)
As shown in Figure \ref{sensor}(a), The omni-directional proximity sensor is an abstraction of sensors that has limited range and resolution in reality, like ultrasonic sensors and radars. The sensor footprint is disk-shaped with a sensing radius of \(d\) and resolution of \(q\). It can perceive the surrounding objects and identify their types with the records from the local database of observed landmarks. The sensor outputs a 3-dimensional matrix \(M_{i \times j \times k}\) as an observation of the agent, in which \(i\), \(j\), \(k\) represent the height, width of the sensor footprint and the type of object in the corresponding cell. The elements in the output matrix outside the sensor footprint are treated as empty cells. Output matrices are used as input tensors of a Convolutional Neural Network (CNN) based encoder for feature extraction. The CNN architecture is shown in Figure \ref{sensor}(b). The tensors passed through 2 convolutional layers and flattened to obtain observation features. The features are finally passed into the Deep Reinforcement Learning (DRL) policy as the input observation for training or inference.

% Rewards
The goal of all agents is to cooperatively construct a landmark complex representation of the environment while adapting for sparse landmarks and reducing client-server communication. Additionally, each agent also needs to avoid colliding with obstacles and other agents while exploring the environment. Therefore, we design a reward function as
$$
r(s_t, a_t) = \sum_{ g \in \mathcal{G}} r_g,
$$
where the set $ \mathcal{G} = \{g: g = s, {comm}, {coll}, {comp}, t \}$ includes five different terms. Since landmarks, or 0-simplices, are nodes when generating the navigation graph from landmark complex representation, more discovered landmarks mean more possible destinations that agents can reach. Thus, agents are expected to discover as many landmarks as possible. Moreover, even if a landmark is already discovered as a node, potential paths to other landmarks, which are edges or 1-simplices, might still remain undiscovered. Therefore, to encourage for constructing the topological representation as complete as possible, agents should receive a higher reward for discovering higher-dimensional simplices. The simplex discovery reward \(r_s\) is expressed as
$$
% r_s = c_{s_0} + 1.5c_{s_1} + 2c_{s_2}
r_s = c_{s_0} + \alpha c_{s_1} + \beta c_{s_2}.
$$
where \(c_{s_0}\), \(c_{s_1}\), and \(c_{s_2}\) denote the number of newly discovered 0-simplices, 1-simplices, and 2-simplices. \(\alpha\) and \(\beta\) are the reward multipliers for 1-simplices and 2-simplices.

In order to reduce client-server communication, an agent is penalized by \(r_{comm}\) whenever it sends a request to the server to exchange information of observed landmarks.
% \[
% r_{comm}=\left\{
%             \begin{array}{ll}
%                 -2 \textit{\indent if communicate with the server,}\\
%                 0 \textit{\indent \:\,  otherwise.}
%             \end{array}
%         \right.
% \]
Prior works in landmark complex exploration \cite{ramaithititima2018landmark, teymouri2021landmark} do not account for collision since they treat robots as particles in numerical simulations. However, as we extend the idea of landmark complex into physics-based environments, collision with both other agents and obstacles should be taken into consideration. When an agent collides with others or obstacles, we apply a penalty \(r_{coll}\) to it.
% \[
% r_{coll}=\left\{
%             \begin{array}{ll}
%                 -5 \textit{\indent if collide with others or obstacles,}\\
%                 0 \textit{\indent \:\,  otherwise.}
%             \end{array}
%         \right.
% \]

To encourage all agents to explore the environment cooperatively, once all landmarks are discovered, the current episode is terminated, and each agent receives a huge group reward \(r_{comp}\) for exploration completion.
% \[
% r_{comp}=\left\{
%             \begin{array}{ll}
%                 5000 \textit{\indent if all landmarks are discovered,}\\
%                 0 \textit{\indent \;\:\:\:\,  otherwise.}
%             \end{array}
%         \right.
% \]
In order to speed up the exploration and further encourage cooperative exploration behavior, each agent receives a small group penalty \(r_t\) at every time step. Combining with \(r_{comp}\), agents should try to fully explore the landmark complex environment as quickly as possible to avoid future time penalty \(r_t\). 
% $$
% r_t = -0.2
% $$

% CL
However, like most modern RL algorithms, MA-POCA suffers from reward sparsity, especially for \(r_{comp}\) that can only be triggered once in each episode. Obstacles and sparse landmarks further amplified this issue. If without any guidance, directly using MA-POCA with the defined reward function results in a poorly converged policy. Therefore, we design a three-stage curriculum to mitigate this problem.

\begin{table}[t]
\caption{Hyperparameters}
\label{hyperparameters}
%\begin{center}
\setlength{\tabcolsep}{17pt}
\begin{tabular}{|c|c|}
\hline
Hyperparameter & Value\\
\hline
Neural network structure & \(2 \times [256, \text{Swish}]\)\\
\hline
Batch size & 1024\\
\hline
Buffer size & 10240\\
\hline
Learning rate & 0.0003\\
\hline
Discount factor & 0.99\\
\hline
Max steps per episode & 20000\\
\hline
\end{tabular}
%\end{center}
\vspace{-0.7cm}
\end{table}

The landmarks in environments of all three stages are placed using LPA. The policy is trained for an equal number of steps in each stage. In stage one, the policy is trained in environments without any obstacles and destroyed landmarks. This stage guides agents to discover the global goal of complete and cooperative exploration by triggering the exploration completion reward \(r_{comp}\) since agents can discover every landmark easily in such simple environments. In stage two, obstacles are generated in the environment randomly and incrementally. Starting with one randomly generated obstacle in the environment, the number of randomly generated obstacles is added by one after every \(e_o\) number of episodes. We model the obstacle as a rectangular with uniformly sampled width $w \in [w_{min}, w_{max}]$ and height $h \in [h_{min}, h_{max}]$. This stage is for agents to adapt more complex environment with decreasing reward signals due to increasing occupied area by obstacles. In addition to random obstacle generation, stage three simulates random landmark destruction. Each landmark placement has a probability of \(p_l\) to fail. \(p_l\) is started with \(5\%\) and incremented by \(5\%\) after every \(n_l\) number of episodes. Once the number of obstacles is incremented, \(p_l\) is reset back to \(5\%\). In this way, agents can experience various combinations of obstacles and landmark destruction possibilities.

\section{Results}
\subsection{Implementation Details}

\begin{figure}[t]
      \centering
      \includegraphics[scale=0.39]{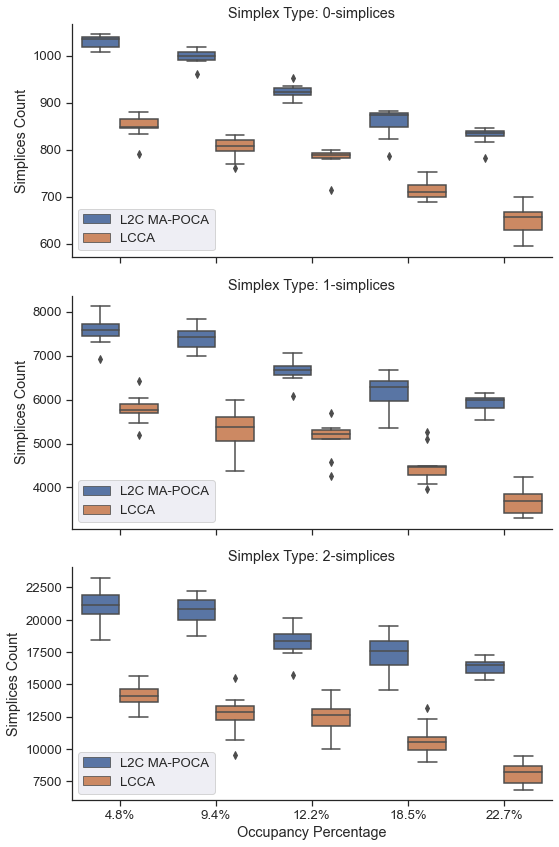}
      \caption{Benchmark tests in environments with various occupancy percentages when the observations count is 1500.}
      \label{benchmark_obstacle}
      \vspace{-0.6cm}
\end{figure}

We use Unity\footnote{https://unity.com/}and ML-Agents\footnote{https://github.com/Unity-Technologies/ml-agents} to simulate physics-based environments and train the DRL policy \cite{juliani2018unity}. The centralized server is developed with Python and hosted in a Docker container using Uvicorn as the application interface. we train the DRL policy on a PC with i7-11800 CPU and RTX 3080 GPU. The hyperparameters of MA-POCA for the training are listed in Table \ref{hyperparameters}.

We use $4$ agents in our implementation. For each agent, we set the sensing resolution $q = 1$m and the sensing radius $d = 15$ m. To better compare with LCCA, we keep the number of agents and range of perception the same as LCCA. The sensor is developed based on the Grid Sensor in Unity Engine, as shown in Figure \ref{sensor} (a). We use a set of sensing radius \(B = \{50, 23, 15\}\) in meters as the sequence of sensor footprints for filtration in LPA. For the reward function, we set the reward multipliers \(\alpha\) as \(1.5\) and \(\beta\) as \(2\). We notice that if \(\alpha\) is set above \(3\) or \(\beta\) is set above \(5\), the policy tends to make considerable communication requests as it can get more rewards from discovering underlying connections between landmarks. We set \(r_{comm}\), \(r_{coll}\), \(r_{comp}\), and \(r_t\) to \(-2\), \(-5\), \(5000\), and \(-0.2\) respectively. For the three-stage curriculum, we set the episode threshold for obstacle increment \(n_o\) as 25 and the episode threshold for incrementation of landmark destruction probability \(n_l\) as 5. The upper bound \(w_{max}\) and lower bound \(w_{min}\) for sampling the width of obstacles are 20 meters and 50 meters, respectively. The upper bound \(h_{max}\) and lower bound \(h_{min}\) for sampling the height of obstacles are 50 meters and 100 meters, respectively. The policy is trained for 5 million steps during each stage.

\begin{figure*}[t]
      \centering
      \includegraphics[scale=0.45]{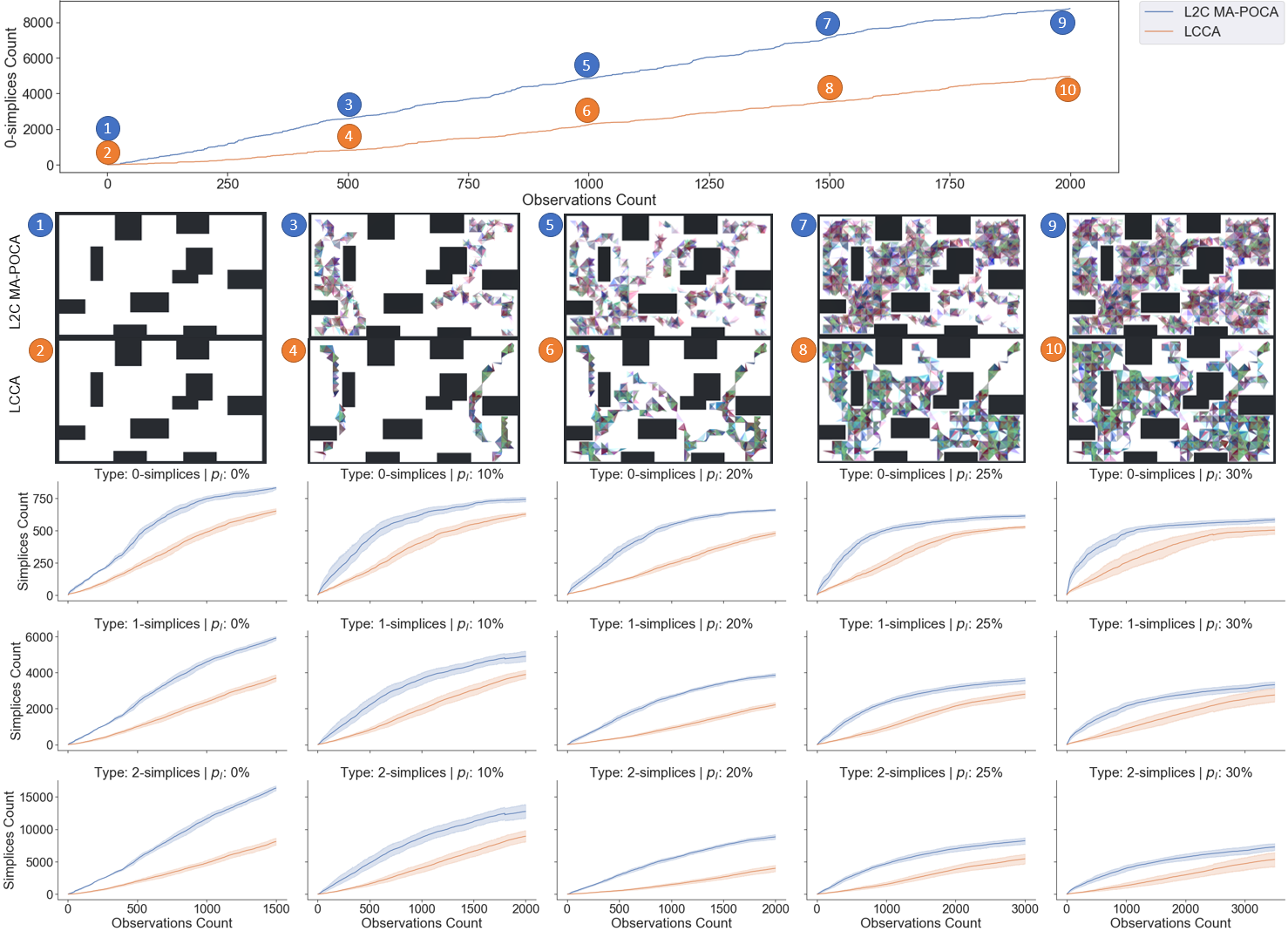}
      \caption{The top plot shows the result of one trial with 20\% landmark destruction probability and 10 obstacles. The screenshots visualize discovered 2-simplices at 5 different moments. The bottom plots show the results of benchmark tests under various landmark destruction probabilities. Means and \(95\%\) confidence intervals are represented by solid lines and colored regions on each plot.}
      \label{benchmark_pl}
      \vspace{-0.55cm}
\end{figure*}

% \begin{figure}[t]
%       \centering
%       \includegraphics[scale=0.35]{training_average_return.png}
%       \caption{Comparisons of the learning curves using different training methods. L2C MA-POCA converges to a considerably better policy compared with standard MA-POCA.}
%       \label{training_average_return}
% \end{figure}

% We use \(v_x \in [-2, 2]\) m/s, \(v_y \in [-2, 2]\) m/s, and \(\omega_z \in [-\frac{5 \pi}{3}, \frac{5 \pi}{3}]\) rad/s as movement action space for training the policy.

% The policy in each stage is trained for 5 million steps. 

\subsection{Benchmark Comparisons}

%%% explain how to compare and why the different dynamics are fair enough. 
To understand the exploration strategy learned by our framework, we conduct experiments in environments with various obstacles and landmark destruction probabilities using the learned policy from L2C MA-POCA and LCCA, which is by far the most refined exploration method for landmark complex environments. We conduct 10 trials using both methods under each condition and collect the data on the server side. Since LCCA is designed for pure numerical simulation, the robots in LCCA are lack of dynamic models.
%the robots in LCCA are not moved by force and do not have the property of velocity
Therefore, directly using time steps as the independent variable cannot reasonably compare our approach with LCCA as the velocity is not controllable in LCCA, and incorporating physics-based dynamics involves heavy modifications of LCCA. To fairly compare the performance between the learned policy and LCCA, we use observations count as the independent variable, which is also used as the independent variable to measure the performance in prior work \cite{teymouri2021landmark}.

% Comparison in environments with various obstacles
The performance of exploration is best measured by the statistics of discovered simplices. The plots in Figure \ref{benchmark_obstacle} show the counts of discovered simplices in environments with various percentages of the area occupied by obstacles and no landmark destroyed when the observations count is 1500 using the learned policy and LCCA. The plots illusrate that agents with the learned policy can adapt to complex environments and overall discover more simplices within the same number of observations compared to LCCA. For example, the mean discovered 0-simplices count is 830 using the learned policy in environments with an occupancy percentage of 22.7\%, whereas the 0-simplices count is 651 when using LCCA under the same condition.

% Comparison in environments with various landmark destruction probability
The plots in Figure \ref{benchmark_pl} show the counts of discovered simplices in environments with various landmark destruction probability and 10 obstacles using the learned policy and LCCA. The mean discovered simplices counts are higher than LCCA in all experiments when using the learned policy, and the discovered simplices counts also reach the plateau earlier than LCCA. Therefore, overall, agents with the learned policy perform exploration more efficiently. Also, compared with the learned policy, the confidence intervals from trials using LCCA become significantly larger when landmark destruction probability increases.

%  We use solid lines and colored regions to represent means and \(95\%\) confidence intervals on each plot. 

\section{Conclusion}
In this work, we proposed L2C MA-POCA, a framework for multi-agent landmark complex exploration in environments with sparse landmarks using MADRL. Our proposed method combines MA-POCA and the curriculum learning framework to efficiently learn a policy that performs cooperative landmark complex exploration using limited range sensors in communication-limited environments. The experiments and analysis indicate that the learning-based exploration method can outperform the traditional model-based methods. The learned policy using L2C MA-POCA has a better performance on efficiency than LCCA, the most refined traditional frontier-based landmark complex exploration method. The proposed framework is a step towards landmark complex exploration in resource-constrained environments and deployment of landmark complex in the real world. Future works will focus on fully decentralized landmark complex exploration and real-world deployment.

\bibliographystyle{plainnat}
\bibliography{references}

\end{document}